\documentclass[a4paper,fleqn]{cas-dc}

\usepackage[authoryear,longnamesfirst]{natbib}
\usepackage{amsmath, amsfonts, amssymb}
\usepackage{hyperref}
\usepackage{url}
\usepackage{booktabs}
\usepackage{multirow} 
\usepackage{amsmath}
\usepackage{graphicx}
\usepackage{cleveref}
\usepackage{tcolorbox}
\usepackage{wrapfig}
\usepackage{colortbl}
\usepackage{tabulary}
\usepackage{amsmath}
\usepackage{amssymb}
\usepackage{amsthm}
\usepackage{microtype}

\newtheorem{definition}{Definition}

\def\tsc#1{\csdef{#1}{\textsc{\lowercase{#1}}\xspace}}
\tsc{WGM}
\tsc{QE}

\begin{document}
\let\WriteBookmarks\relax
\def\floatpagepagefraction{1}
\def\textpagefraction{.001}

\shorttitle{Breaking the Tie}    


\title [mode = title]{Breaking the Tie: A Cluster-Aware Routing Framework for Large Language Models}  



\author[label1,label2]{Yao Lu}
\credit{Conceptualization, Methodology, Software, Writing - original draft, Visualization}
\author[label1,label2]{Zhaiyuan Ji}
\credit{Conceptualization, Methodology, Writing - review \& editing}
\author[label1,label2]{Yaxin Gao}
\credit{Validation, Formal analysis, Investigation}
\author[label1,label2]{Zeyu Wang}
\credit{Formal analysis, Investigation}
\author[label2]{Zhe Tang}
\credit{Investigation, Data curation}
\author[label3]{Jiaheng Wei}
\credit{Investigation, Visualization}
\author[label2]{Zhaowei Zhu}
\credit{Visualization, Writing - review \& editing, Supervision}
\author[label1,label2]{Shanqing Yu}
\cormark[1]
\credit{Resources, Writing - review \& editing, Supervision}
\author[label1,label2]{Qi Xuan}
\credit{Resources, Funding acquisition}

\cortext[cor1]{Corresponding author: Shanqing Yu (yushanqing@zjut.edu.cn).}
\affiliation[label1]{organization={Institute of Cyberspace Security, Zhejiang University of Technology}}

\affiliation[label2]{organization={Binjiang Institute of Artificial Intelligence, Zhejiang University of Technology}}

\affiliation[label3]{organization={Hong Kong University of Science and Technology (Guangzhou)}}

\fntext[1]{This work was supported by Key Technology Research and Development Program Project of Hangzhou (2025SZD1A41), the Baima Lake Laboratory Joint Fund of the Zhejiang Provincial Natural Science Foundation of China (LBMHZ25F020002), the National Key R\&D Program of China (2025YFA1510900), the Zhejiang Provincial Key R\&D Program (2024C01025) and the National Natural Science Foundation of China (62503423).}


\begin{abstract}
With the rapid development of artificial intelligence, the emergence of various Large Language Models (LLMs) has created a rich model ecosystem. However, this also brings a key challenge: how to select the optimal model for a specific user query. LLM routing addresses this need by dynamically assigning queries to the most suitable expert in the pool of candidate models. However, existing routing frameworks often simplify this process to a standard classification task; thus, a critical vulnerability is exposed when multiple candidate models correctly answer the same query. We formalize this capability overlap as routing noise, which misleads the router with arbitrarily correct candidate models, ultimately leading to routing collapse (a severe decline in generalization ability on unseen tasks). To address this problem, we propose a novel \textbf{C}luster-\textbf{A}ware \textbf{S}oft-\textbf{L}abeling \textbf{R}outing (\textbf{CASLR}) framework. CASLR shifts the evaluation paradigm from the success of a single query to macro-domain consensus by replacing traditional one-hot vectors with a masked softmax mechanism. Specifically, for experts who answer incorrectly, we penalize their target probability to zero; for the remaining candidates, we directly compute continuous fine-grained soft labels based on their global clustering utility scores. We then use these refined soft labels to supervise a lightweight router. Specifically, the framework not only demonstrates superior accuracy on multiple benchmarks, but also outperforms Llama-3.3-70B-Instruct by $7.80\%$ in overall average performance. Furthermore, the extremely low routing inference latency of only $1.13$s further confirms that CASLR can achieve efficient system scheduling with almost zero additional overhead, while ensuring high response quality.
\end{abstract}


\begin{highlights}
\item We define Routing Noise and Routing Collapse that exist in LLM routing methods.
\item We propose the CASLR framework for high-precision expert dispatching.
\item Extensive experiments demonstrate that CASLR outperforms SoTA LLM routing baselines.
\end{highlights}



\maketitle

\section{Introduction}
In recent years, LLMs have demonstrated remarkable capabilities in natural language understanding~\citep{liu2019multi,xiao2026reversible}, machine translation~\citep{wu2016google}, code generation~\citep{li2022competition}, and complex reasoning tasks~\citep{lu2024reassessing,lu2025autoannotator,gao6810312mapping,lu2025structural,zhou2026comem}. With the gradual prosperity of the open source community, the number of available models has grown exponentially\footnote{As of March 20, 2026, the Huggingface platform alone has 2,722,255 models.}, forming an extremely rich model ecosystem, from general-purpose models with trillions of parameters to specialized models optimized for specific domains. However, despite the abundance of models, no single model can maintain optimal performance across all tasks~\citep{hu2024routerbench,stripelis2024tensoropera}, posing a challenge for users in identifying the model best suited for their applications.

An early solution is the Mixture of Experts~\citep{cai2024survey,riquelme2021scaling} (MoE), which integrates multiple specialized sub-networks (experts) within a model and dynamically activates the most relevant expert for each input through a gating network. The MoE paradigm enables models to achieve higher performance than dense models with similar computational costs and handle diverse inputs more effectively. However, these models have inherent limitations: The expert network is tightly integrated into a single model, making it difficult to flexibly replace or upgrade to heterogeneous experts as technology evolves.

\begin{figure}[pos=t]
    \centering
    \includegraphics[width=0.45\textwidth]{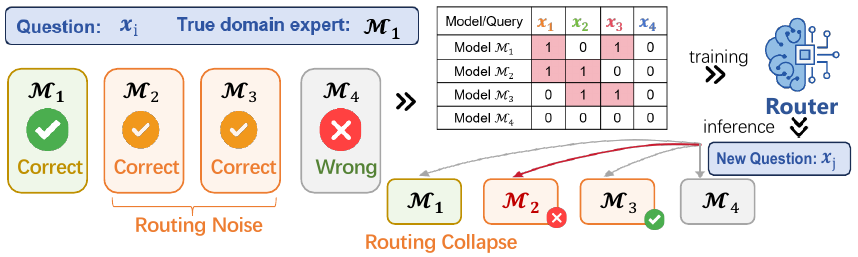}
    \caption{Visualization of routing noise and routing collapse.}
    \label{fig:issue}
\end{figure}

To solve this problem, LLM routing~\citep{hu2024routerbench,chen2023frugalgpt} has emerged as an effective method for fully utilizing heterogeneous resources, gaining significant attention from academia and industry. The core idea of LLM routing is to dynamically select the "most suitable" model to execute the task from a candidate pool containing multiple heterogeneous LLMs for each input query. For example, FrugalGPT~\citep{chen2023frugalgpt} utilizes a cascading strategy to query the model sequentially, thereby significantly reducing inference costs while maintaining generation quality. BEST-Route~\citep{ding2025best} uses a multi-head router based on DeBERTa-v3-small~\citep{he2020deberta} to estimate query difficulty and select the optimal model. However, despite the promising progress made by existing routing frameworks, most of them simplify this dynamic routing problem to a simple classification~\citep{huang2025routereval} or similarity matching problem. This exposes a key flaw: these frameworks perform poorly in scenarios where model capabilities are commensurate—that is, when multiple candidate experts can correctly answer the same input query. Faced with these overlapping capabilities, traditional routing mechanisms are often severely affected by \textbf{\textit{routing noise}}, as illustrated in \Cref{fig:example}. They cannot effectively distinguish between a model that happens to give the correct answer and a true expert in the field. Due to the lack of an effective tie-breaking mechanism, routers are highly susceptible to absorbing these false signals during training, leading to severe \textbf{\textit{routing collapse}}.

To address this issue, we propose a novel \textbf{C}luster-\textbf{A}ware \textbf{S}oft-\textbf{L}abeling \textbf{R}outing (\textbf{CASLR}) framework. This framework comprises two core phases: offline cluster-aware dataset construction as well as router training and inference. During the dataset construction phase, we shift the evaluation paradigm from single-query success to macro-domain consensus. Specifically, we project queries into a continuous embedding space and apply unsupervised clustering to group them into different semantic task domains. Then, inspired by logit-based knowledge distillation~\citep{sun2024logit,gao2025maintaining}, we replace one-hot vectors with a masked softmax mechanism to generate fine-grained soft labels: for experts who answer the query incorrectly, we explicitly penalize their target probability to zero; for the remaining candidate experts, we compute their continuous soft labels by directly applying the softmax function to their global clustering utility scores. By pairing the original queries with these computed soft labels, we formally construct a high-quality routing training dataset. In the subsequent router training and inference phase, we utilize this newly constructed dataset to supervise a lightweight bidirectional encoder (e.g., BERT~\citep{devlin2019bert}). By optimizing the router against these refined soft labels rather than one-hot labels, CASLR fundamentally mitigates the routing collapse phenomenon induced by routing noise, ensuring highly accurate expert matching.

In summary, our main contributions are as follows:
\begin{itemize}
    \item We formally define \textbf{Routing Noise} and \textbf{Routing Collapse}, providing a rigorous theoretical perspective that reveals the fatal flaws of traditional hard-label routing strategies when faced with overlapping capabilities of multiple models.
    \item To tackle this routing collapse problem, we propose a novel \textbf{C}luster-\textbf{A}ware \textbf{S}oft-\textbf{L}abeling \textbf{R}outing (\textbf{CASLR}) framework. We first introduce a masked soft-labeling mechanism based on macro-domain consensus to filter out incidental routing noise. Then, we utilize these refined soft labels to supervise a lightweight BERT router, enabling it to achieve high-precision expert scheduling with extremely low latency.
    \item Extensive experiments across various datasets demonstrate that CASLR significantly outperforms competitive baselines, including both standard Mixture-of-Experts models and dedicated LLM routing methodologies.
\end{itemize}

\section{Related Work}
\label{sec:Related Work}

\textbf{Mixture-of-Expert.} MoE, as a classic conditional computation paradigm, is based on the idea of dynamically allocating input samples to specific expert subnetworks through gating networks, thereby significantly expanding the model capacity without significantly increasing the amount of inference computation. In 2020, GShard~\citep{lepikhin2020gshard} introduces MoE technology into the Transformer architecture for the first time, enabling MoE models to be trained and inferred on a large scale in a distributed environment, further promoting the application of MoE models in the field of natural language processing. Subsequently, Switch Transformer~\citep{fedus2022switch} and GLaM~\citep{du2022glam} further explore the application potential of MoE technology in natural language processing, achieving superior performance through optimized gating mechanisms and expert design. In recent years, advanced open-source LLMs have further promoted the refined evolution of MoE mechanisms. For example, Mixtral 8x7B~\citep{jiang2024mixtral} builds upon the foundational Mistral 7B~\citep{jiang2023mistral7b} architecture by scaling each layer into a sparse mixture of eight FFNs. DeepSeekMoE~\citep{dai2024deepseekmoe} introduces the concept of fine-grained expert segmentation by subdividing the intermediate hidden dimension of the FFN expert, while preserving the overall parameter count.

\textbf{LLM Routing.} With the flourishing development of LLMs, the differences in cost, latency, and domain capabilities among different models have become increasingly prominent. This has directly spurred the rise of LLM routing technology, which aims to dynamically distribute user requests to the most suitable model. In existing related studies, researchers have explored different distribution strategies from different perspectives: for example, Frugal-GPT~\citep{chen2023frugalgpt} follows a cascading approach, sequentially querying LLMs until a reliable response is obtained. GraphRouter~\citep{feng2024graphrouter} frames LLM selection as inductive edge prediction over a task–query–model graph, enabling cost-performance estimation and effortless integration of new LLMs. To comprehensively benchmark this rapidly growing body of algorithms, RouterBench~\citep{hu2024routerbench} introduces a comprehensive dataset of 405k multi-inference results from representative LLMs to support the development of routing strategies. Subsequently, BEST-Route~\citep{ding2025best} uses a multi-head router based on DeBERTa-v3-small~\citep{he2020deberta} to estimate query difficulty and select the optimal model. Unlike other methods that treat the router merely as a "Query Distributor", the core innovation of Router-R1~\citep{zhang2025router} lies in enabling the router itself to become a Policy LLM with reasoning capabilities.

\section{Preliminary}
\textbf{Pipeline of LLM Routing.} Given a model zoo $\mathcal{P}=\{\mathcal{M}_1,\mathcal{M}_2,\cdots,\mathcal{M}_N\}$ composed of $N$ well-trained LLMs, model routing involves selecting an optimal model $\mathcal{M}^{*}$ from $\mathcal{P}$ for each user query $x \in \mathcal{X}$, where $\mathcal{X}$ denotes the space of all possible user inputs. The selection is typically performed by a routing function $f: \mathcal{X} \rightarrow \mathcal{P}$, designed to maximize a specific expected utility function $\mathcal{U}(x, \mathcal{M})$ (e.g., response quality or domain matching score). Formally, the routing process is defined as:
\begin{equation}
\mathcal{M}^{*} = f(x) = \underset{\mathcal{M}_k \in \mathcal{P}}{\arg\max} \ \mathcal{U}(x, \mathcal{M}_k).
\end{equation}


\section{Method}
\begin{figure*}[pos=t]
    \centering
    \includegraphics[width=0.95\linewidth]{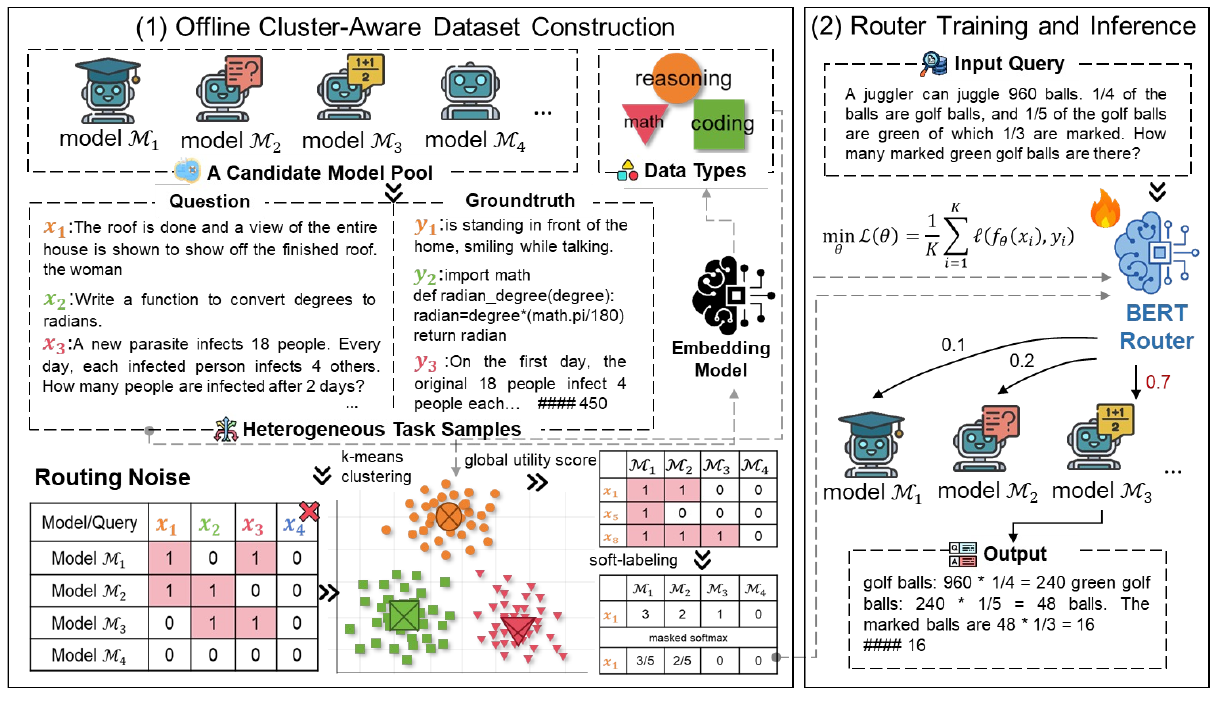}
    \caption{Overview of the CASLR framework.}
    \label{fig:all framework}
\end{figure*}

The core objective of our framework is to dynamically route a user query $x$ to the most suitable LLM from a pool of $N$ candidate experts, $\mathcal{P}=\{\mathcal{M}_1,\mathcal{M}_2,\cdots,\mathcal{M}_N\}$. As illustrated in \Cref{fig:all framework}, our framework can be divided into an offline cluster-aware dataset construction phase and a router training and inference phase.

\subsection{Offline Cluster-aware Dataset Construction}
The foundation of our routing framework lies in building a high-quality, domain-aware dataset $\mathcal{D} = \{(x_i, y_i)\}|_{i=1}^{K}$ for routing training, where $x_i$ is the query and $y_i \in \{1, 2, \dots, N\}$ is the optimal expert index. For each query $x_i$, we first evaluate all $N$ candidate models to obtain a correctness vector $\mathbf{v}_i \in \{0, 1\}^N$, where the $i$-th element indicates whether expert $\mathcal{M}_i$ answered correctly. However, when multiple experts give the correct answer (i.e., $\sum \mathbf{v}_i > 1$), routers will be unable to distinguish between genuine domain experts and inferior models that happen to give the correct answer. We formally define this phenomenon as \textbf{Routing Noise} and give its complete definition below:
\begin{definition}[Routing Noise]
\label{def:routing_noise}
For a given query $x_i$ and its true domain expert $\mathcal{M}^*$, \textbf{Routing Noise} is defined as the subset of candidate models that achieve the same performance as $\mathcal{M}^*$ on $x_i$, but possess inferior intrinsic capability in the broader domain $\mathcal{D}_{x_i}$. Mathematically, the routing noise set $\mathcal{N}(x_i)$ can be formulated as:
\begin{equation}
\label{eq:routing_noise}
\begin{aligned}
    \mathcal{N}(x_i) = \Big\{ \mathcal{M}_k \in \mathcal{P} & \setminus \{\mathcal{M}^*\} \mid {}  \mathcal{U}(\mathcal{M}_k, x_i) = \mathcal{U}(\mathcal{M}^*, x_i) \\
    & \text{ and } \mathcal{U}(\mathcal{M}_k, \mathcal{D}_{x_i}) < \mathcal{U}(\mathcal{M}^*, \mathcal{D}_{x_i}) \Big\}.
\end{aligned}
\end{equation}
\end{definition}
Due to this routing noise, the router is highly susceptible to absorbing false signals during training, severely degrading its generalization capability. We formalize this systemic failure as \textbf{Routing Collapse} and define it as follows:
\begin{definition}[Routing Collapse]
\label{def:routing_collapse}
Given a query $x_i$, \textbf{Routing Collapse} is defined as the systemic performance degradation caused by the router's inability to distinguish the true domain expert $\mathcal{M}^*$ from the routing noise $\mathcal{N}(x_i)$ using a routing policy $\pi_\theta$.
\begin{equation}
\label{eq:routing_collapse}
    \begin{aligned}
    \Delta \mathcal{U}(x_i) = {} & \mathcal{U}(\mathcal{M}^*, \mathcal{D}_{x_i}) \\
    & {} - \sum_{\mathcal{M}_k \in \mathcal{N}(x_i)} \pi_\theta(\mathcal{M}_k \mid x_i) \\
    & \qquad {} \cdot \mathcal{U}(\mathcal{M}_k, \mathcal{D}_{x_i}).
\end{aligned}
\end{equation}
\end{definition}
To fundamentally avoid routing collapse, we introduce a cluster-aware expert ranking mechanism that identifies the optimal domain expert through global domain consensus. Specifically, we first map each query into a high-dimensional semantic space using a pre-trained embedding model $\phi(\cdot)$:
\begin{equation}
    z_i = \phi(x_i), \quad z_i \in \mathbb{R}^d.
\end{equation}
Given the whole set of embedding vectors $\mathcal{Z} = \{z_i\}_{i=1}^{K}$, we apply k-means clustering~\citep{hartigan1979algorithm} to partition the semantic space into $n$ disjoint clusters $\mathcal{C} = \{C_1, C_2, \dots, C_n\}$. These clusters essentially group queries representing similar tasks and semantic structures together. We use these clustered queries as fine-grained evaluation signals to further filter out the truly optimal expert models. The basic logic is simple: a poorly performing model, if it correctly answers a single query only by chance (thus becoming a source of routing noise), is highly unlikely to maintain its accuracy across a variety of semantically related queries. Conversely, true domain experts will consistently exhibit robust performance across all queries in the same cluster. Guided by this rationale, we shift from single-query evaluation to cluster-level consensus to resolve routing collapse during training dataset construction. Specifically, for a query $x_i$ belonging to cluster $C_i$, we compute a global utility score $\mathcal{U}_{g}(\mathcal{M}_k, C_i)$ for each candidate expert $\mathcal{M}_k$ who answers correctly on $x_i$:
\begin{equation}
\mathcal{U}_{g}(\mathcal{M}_k, C_i) = \sum_{x_i \in C_i} v_{i,k}.
\end{equation}
After obtaining the global utility score $\mathcal{U}_{g}$, a naive way to construct the labels for the routing dataset $\mathcal{D}$ is to assign a fixed one-hot label by selecting an expert who can both correctly answer the query $x_i \in C_i$ and achieve the highest performance in clusters $C_i$:
\begin{equation}
y_i = \text{argmax}_{\mathcal{M}_k} \{ \mathcal{U}_{g}(\mathcal{M}_k, C_i) \mid v_{i,k} = 1, x_i \in C_i \}.
\end{equation}
However, this hard-labeling strategy is inherently suboptimal. It enforces a strict "winner-takes-all" assignment through sparse one-hot vectors, incorrectly treating candidate experts as mutually exclusive. Besides, this hard-labeling strategy forces the router to over-rely on a single expert, resulting in overly sharp decision boundaries, severely exacerbating the risk of overfitting and reducing the model's generalization ability to unseen queries.

To overcome these limitations, we propose an improved soft-labeling mechanism inspired by logit-based knowledge distillation~\citep{sun2024logit,gao2025maintaining}. Specifically, for experts who fail to answer query $x_i$, we explicitly penalize their target probability to zero. For all candidates who answer query $x_i$ correctly, we normalize their global utility scores using a masked softmax operation:
\begin{equation}
\label{eq:soft_label}
    y_{i,k} = 
    \begin{cases} 
    \frac{\exp(\mathcal{U}_{g}(\mathcal{M}_k, C_i))}{\sum_{\mathcal{M}_m \in (\mathcal{M}^* \bigcup \mathcal{N}(x_i))} \exp(\mathcal{U}_{g}(\mathcal{M}_m, C_i))}, & \substack{\text{if } \mathcal{M}_k \in \\ (\mathcal{M}^* \bigcup \mathcal{N}(x_i))} \\
    0, & \text{otherwise}
    \end{cases}
\end{equation}
Once the optimal routing label $\mathcal{D} = \{(x_i, y_i)\}_{i=1}^N$ is determined, we can formulate LLM routing as minimizing the cross-entropy loss between its prediction distribution and the refined soft label.

\subsection{Router Training and Inference}
With the refined training dataset $\mathcal{D} = \{(x_i, y_i)\}_{i=1}^K$ established, we formulate the expert selection process as a supervised classification task. To learn the mapping from query semantics to the refined expert labels, we employ a bidirectional encoder, specifically BERT~\citep{devlin2019bert}, as our router. We adopt BERT~\citep{devlin2019bert} rather than generative LLMs for two primary reasons:
\begin{itemize}
    \item \textbf{Minimal Inference Overhead}: BERT introduces negligible parameter overhead and computational latency compared to using a decoder-only LLM as a router, ensuring the overall system remains highly efficient.
    \item \textbf{Competitive Classification Capability}: Despite having a small number of parameters, encoder-only models like BERT are not inferior to LLMs in sequence classification.
\end{itemize}
Driven by these advantages, we formalize the router's training objective as the minimization of the cross-entropy loss over the refined dataset:
\begin{equation}
\label{eq:loss_function}
    \min_\theta \mathcal{L}(\theta) = \frac{1}{K} \sum_{i=1}^K \ell \left( f_\theta(x_i), y_i \right)
\end{equation}
where $\theta$ represents the trainable parameters of the BERT $f_{\theta}$ and $f_\theta(x_i)$ denotes the model's prediction of assigning query $x_i$.

During the inference phase, an unseen query $x_{test}$ is fed into the trained router to obtain a probability distribution over the expert pool. The system then routes the query to the expert with the maximum probability:
\begin{equation}
\label{eq:inference_dispatch}
    \mathcal{M}^* = \arg\max f_\theta(x_{test})
\end{equation}
Finally, the selected expert $\mathcal{M}^*$ performs the final inference.

\section{Experiments}
\subsection{Baseline Methods}
To demonstrate the effectiveness of our method, we compare it with the following baselines: Oracle~\citep{srivatsa2024harnessing}, Random~\citep{srivatsa2024harnessing}, Random\_Oracle\_0.5~\citep{hu2024routerbench}, select the best expert model,
LinearR~\citep{huang2025routereval}, MLPR~\citep{huang2025routereval}, C-Bert~\citep{huang2025routereval}, MLC~\citep{huang2025routereval}, PRknn~\citep{huang2025routereval} and GraphRouter~\citep{feng2024graphrouter}. We provide the detailed descriptions of these methods in \Cref{sec:Introduction to Baseline Methods}.

\subsection{Models}
In this paper, we utilize several advanced open-source LLMs to construct a candidate expert database. Specifically, the selected models include: Qwen2.5-7B-Instruct~\citep{qwen2.5}, Qwen2.5-Coder-7B-Instruct~\citep{qwen2.5}, Qwen2.5-Math-7B-Instruct~\citep{yang2024qwen25mathtechnicalreportmathematical}, Deep-Seek-R1-Distill-Llama-8B~\citep{deepseekai2025deepseekr1incentivizingreasoningcapability}, DeepSeek-R1-Distill-Qwen-14B~\citep{deepseekai2025deepseekr1incentivizingreasoningcapability}, Llama-3.1-8B-Instruct\footnote{https://huggingface.co/meta-llama/Llama-3.1-8B-Instruct}, Llama-3.3-70B-Instruct\footnote{https://huggingface.co/meta-llama/Llama-3.3-70B-Instruct}, Qwen3-30B-A3B~\citep{qwen3technicalreport}, Qwen3-30B-A3B-Instruct-2507~\citep{qwen3technicalreport}, Qwen3-Coder-30B-A3B-Instruct~\citep{qwen3technicalreport} and Mixtral-8x7B-Instruct-v0.1~\citep{jiang2024mixtral}. To further test the router's fine-grained scheduling capabilities, we specifically introduce Qwen2.5-Coder-7B-Instruct and Qwen2.5-Math-7B-Instruct into the candidate pool as "domain specialists". Meanwhile, the remaining models serve as "general models", responsible for broadly covering various complex open-domain problems.

\subsection{Evaluation and Datasets} 
\label{sec:Evaluation and Datasets}
To validate the performance of our routing framework, we construct a comprehensive evaluation system comprising 7 core tasks, covering 3 dimensions: mathematical computation, code generation, and logical reasoning. Specifically, we use MATH~\citep{hendrycks2021measuring}, GSM-Symbolic~\citep{mirzadeh2024gsm} and AIME1983-2025~\citep{aime_1983_2024,aime_2025} to validate mathematical problem-solving capabilities; HumanEval~\citep{chen2021evaluating} and MBPP~\citep{austin2021program} are used to evaluate code generation capabilities; and Hellaswag~\citep{zellers2019hellaswag} and MMLU~\citep{hendrycks2020measuring} are introduced to capture broad world knowledge and common-sense reasoning abilities. All experiments are conducted on 4 NVIDIA A800 80G GPUs.

\begin{table*}[pos=t]
  \centering
     \caption{Performance comparison with state-of-the-art LLM routing methods. We select the first $5$ models in \Cref{tab:base model acc} as experts.}\label{tab:sota}%
   \resizebox{\textwidth}{!}{
    \begin{tabular}{c|ccc|cc|cc|c}
    \toprule
    Task  & \multicolumn{3}{c|}{Math} & \multicolumn{2}{c|}{Coding} & \multicolumn{2}{c|}{Reasoning} & \multirow{2}[4]{*}{Average} \\
\cmidrule{1-8}    Models\textbackslash{}Dataset & MATH  & GSM-Symbolic & AIME1983-2025  & HumanEval & Mbpp  & MMLU  & HellaSwag &  \\
    \midrule
    Oracle & \textbf{89.40\%} & \textbf{97.10\%} & \textbf{78.06\%} & \textbf{93.94\%}& \textbf{89.11\%} & \textbf{93.04\%}  & \textbf{89.36\%} & \textbf{90.00\%} \\
    Random & 70.12\% & 81.11\% & 39.99\% & 70.35\% & 60.96\% & 69.95\%  & 58.17\% & 64.38\% \\
    Random\_Oracle\_0.5 & 79.79\% & 89.17\% & 59.07\% & 82.33\% & \underline{75.16\%} & 81.43\%  & \underline{73.76\%} & 77.24\% \\
    PRknn & 76.00\% & 91.90\% & 67.35\% & 78.79\% & 68.00\% & 76.36\%  & 65.90\% & 74.90\% \\
    LinearR & 79.76\% & 91.30\% & 60.71\% & \underline{90.91\%} & 72.00\% & 82.38\%  & 69.85\% & 78.13\% \\
    MLPR  & 77.52\% & 92.70\% & 62.24\% & 84.85\% & 72.00\% & 77.32\%  & 65.20\% & 75.98\% \\
    C-Bert & 70.29\% & 88.20\% & 23.60\% & 66.67\% & 54.08\% & 73.37\%  & 67.53\% & 63.39\% \\
    MLC & 78.98\% & 92.68\% & 73.03\% & 77.78\% & 65.31\% & 79.03\%  & 66.19\% & 76.14\% \\
    GraphRouter & 66.66\% & 88.74\% & 23.80\% & 85.98\% & 57.31\% & 74.16\% & 59.11\% & 65.11\% \\
    Select the best expert model & 76.92\% & 89.80\% & \underline{68.37\%} & \textbf{93.94\%} & 68.00\% & \underline{82.66\%}  & 72.05\% & 78.82\% \\
    \cellcolor[rgb]{ .867,  .922,  .969}Ours  & \cellcolor[rgb]{ .867,  .922,  .969}\underline{81.44\%} & \cellcolor[rgb]{ .867,  .922,  .969}\underline{93.20\%} & \cellcolor[rgb]{ .867,  .922,  .969}66.33\% & \cellcolor[rgb]{ .867,  .922,  .969}\textbf{93.94\%} & \cellcolor[rgb]{ .867,  .922,  .969}69.00\% & \cellcolor[rgb]{ .867,  .922,  .969}82.52\%  & \cellcolor[rgb]{ .867,  .922,  .969}72.10\% & \cellcolor[rgb]{ .867,  .922,  .969}\underline{79.79\%} \\
    \bottomrule
    \end{tabular}}
\end{table*}%

\begin{figure}[pos=htbp]
    \centering
    \includegraphics[width=0.49\textwidth]{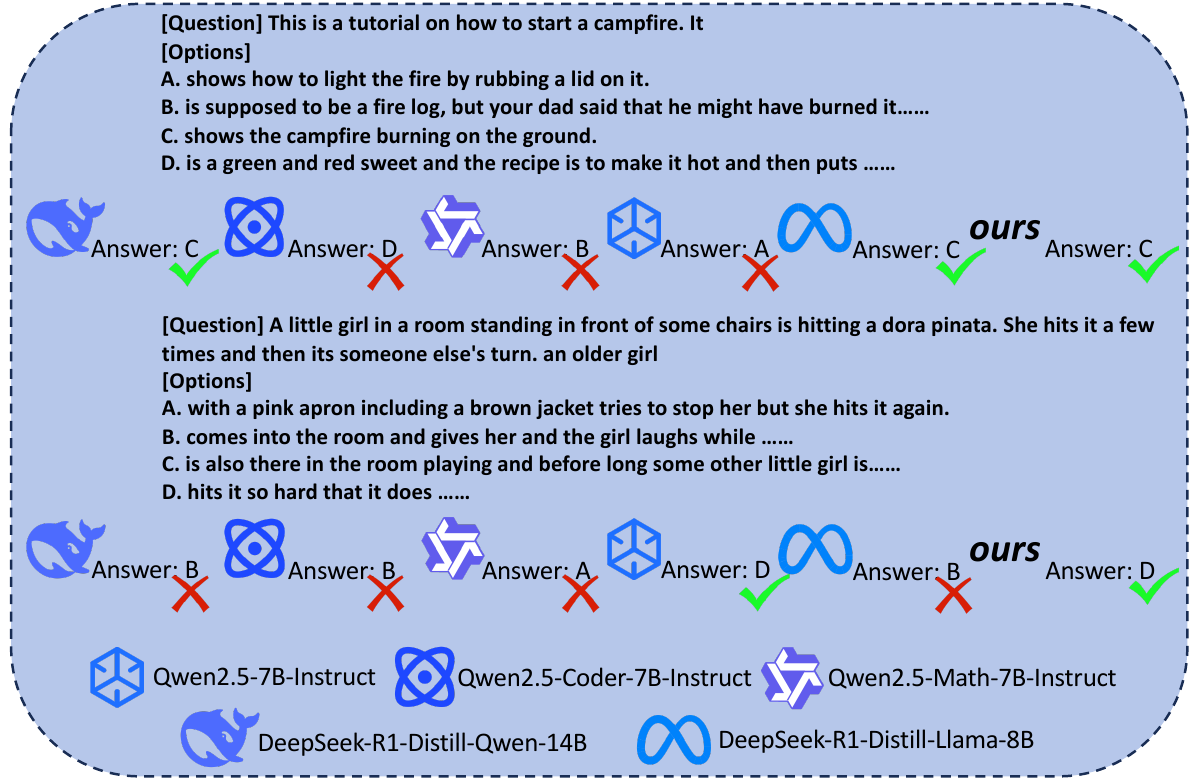}
    \caption{Case studies of model predictions: Ours vs. existing LLM models.}
    \label{fig:example}
\end{figure}

\begin{table*}[pos=t]
  \centering
  \caption{Performance comparison with large-scale LLM models. Experiments are conducted using the 5 best-performing models.}\label{tab:scale to large model}%
  \resizebox{\textwidth}{!}{
    \begin{tabular}{c|ccc|cc|cc|c}
    \toprule
    Task  & \multicolumn{3}{c|}{Math} & \multicolumn{2}{c|}{Coding} & \multicolumn{2}{c|}{Reasoning} & \multirow{2}[4]{*}{Average} \\
\cmidrule{1-8}    Models\textbackslash{}Dataset & MATH  & GSM-Symbolic & AIME1983-2025 & HumanEval & Mbpp  & MMLU   & HellaSwag &  \\
    \midrule
    Qwen3-30B-A3B & 72.76\% & 74.20\% & 79.59\% & 69.70\% & 71.00\% & 85.29\% & 74.75\% & 75.33\% \\
    Qwen3-30B-A3B-Instruct-2507 & 83.24\% & 76.40\% & 81.63\% & \underline{90.91\%} & 71.00\% & 85.18\%  & 78.15\% & 80.93\% \\
    Mixtral-8x7B-Instruct-v0.1 & 23.32\% & 62.80\% & 1.53\% & 54.55\% & 41.00\% & 68.89\%  & 63.10\% & 45.03\% \\
    Qwen3-Coder-30B-A3B-Instruct & 82.20\% & 91.10\% & 49.49\% & \underline{90.91\%} & 71.00\% & 82.05\%  & 71.40\% & 76.88\% \\
    Llama-2-13B & 3.64\% & 26.00\% & 0.51\% & 9.09\% & 22.00\% & 47.48\%  & 52.55\% & 23.04\% \\
    Qwen2.5-Coder-32B-Instruct & 73.56\% & 86.80\% & 31.63\% & 75.76\% & \underline{76.00\%} & 78.41\%  & 71.85\% & 70.57\% \\
    Llama-3.3-70B-Instruct & 80.60\% & 93.50\% & 55.10\% & 81.82\% & 64.00\% & 86.30\%  & 79.25\% & 77.22\% \\
      Oracle (Best 5 models)  & \textbf{93.86\%} & \textbf{97.64\%} & \textbf{92.65\%} & \textbf{96.95\%} & \textbf{84.00\%} & \textbf{95.11\%} & \textbf{92.25\%} & \textbf{93.21\%} \\
    \cellcolor[rgb]{ .867,  .922,  .969}Ours (Best 5 models)  & \cellcolor[rgb]{ .867,  .922,  .969}\underline{84.80\%} & \cellcolor[rgb]{ .867,  .922,  .969}\underline{96.20\%} & \cellcolor[rgb]{ .867,  .922,  .969}\underline{82.14\%} & \cellcolor[rgb]{ .867,  .922,  .969}\underline{90.91\%} & \cellcolor[rgb]{ .867,  .922,  .969}72.00\% & \cellcolor[rgb]{ .867,  .922,  .969}\underline{88.64\%}  & \cellcolor[rgb]{ .867,  .922,  .969}\underline{80.45\%} & \cellcolor[rgb]{ .867,  .922,  .969}\underline{85.02\%} \\
    \bottomrule
    \end{tabular}}
\end{table*}%

\subsection{Main Experiments}
\textbf{Comparison with State-of-the-Art Routing Methods.} To demonstrate the effectiveness of our proposed method, we compared it with a range of state-of-the-art routing baselines. Herein, we select the first $5$ models in \Cref{tab:base model acc} as experts for the experiment. These baselines include heuristic strategies (Random, Oracle, Random\_Oracle\_0.5, select the best expert), classical machine learning methods (PRknn, C-Bert) and advanced neural routing frameworks (LinearR, MLC, MLPR, GraphRouter). It is worth noting that Oracle represents the theoretical upper bound where a query is marked correct if any model in the candidate pool answers it successfully. As illustrated in \Cref{tab:sota}, our proposed method establishes a new state-of-the-art among all non-oracle routing baselines, achieving an average accuracy of $79.79\%$. Among all baseline methods, the second-best performing is the heuristic strategy "Select the best expert model". Our method outperforms it by an absolute $0.97\%$. This demonstrates that relying solely on a single expert who performs best across the entire dataset is insufficient. Our routing framework successfully achieves dynamic distribution, fully activating the complementary strengths of multiple experts, rather than getting bogged down in the performance bottleneck of a single model. Besides, it significantly outperforms previous LLM routing methods, achieving a remarkable $14.68\%$ higher score than GraphRouter and $3.65\%$ higher than MLC. These empirical results confirm that our cluster-aware soft-labeling mechanism endows the router with acute domain awareness. To visually demonstrate the superior performance of our CASLR, \Cref{fig:example} provides two specific examples based on common-sense reasoning. The results show that CASLR consistently avoids distracting options and accurately assigns queries to genuine domain experts, thus ensuring the correctness of the final results.


\begin{figure}[pos=t]
    \centering
    \includegraphics[width=0.49\textwidth]{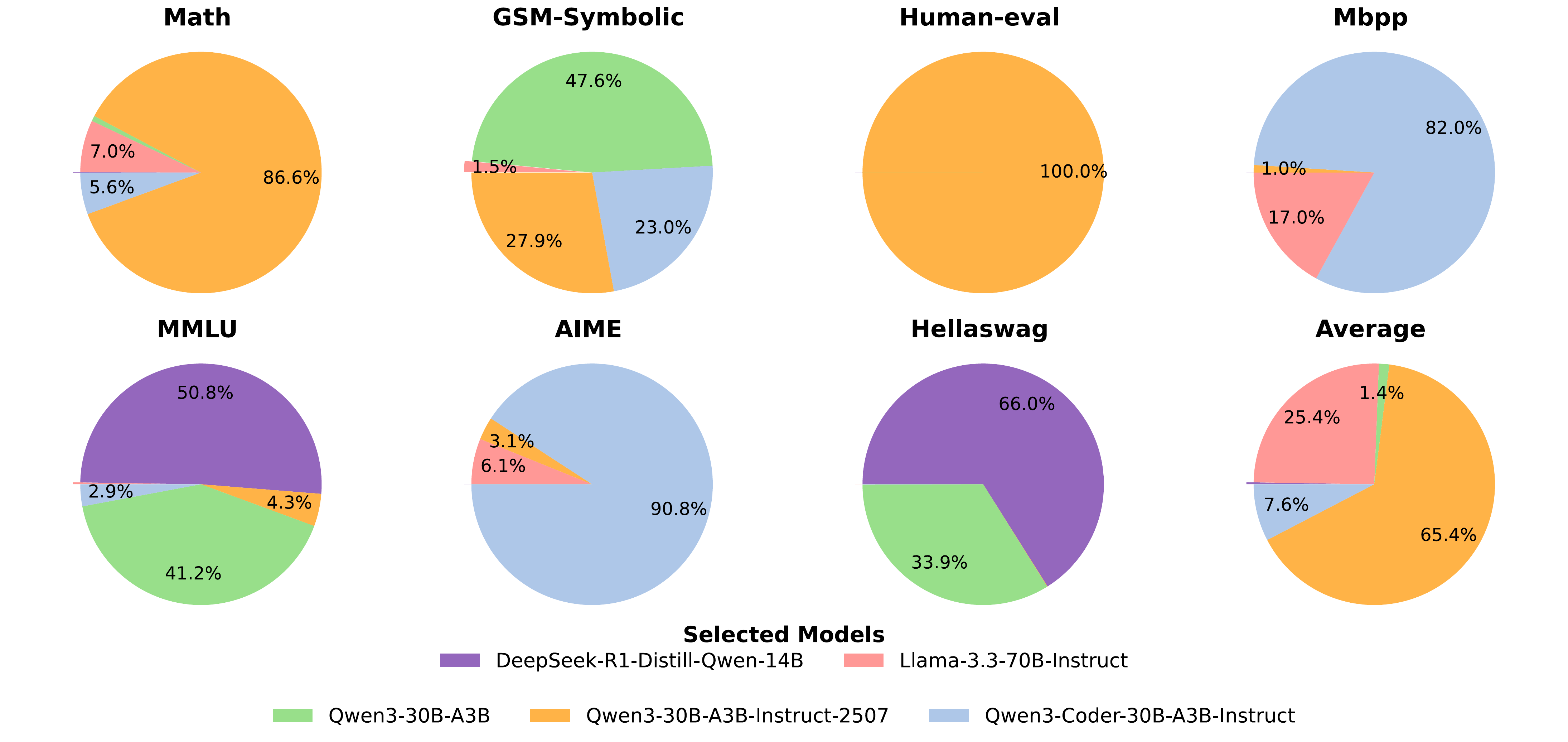}
    \caption{Percentage of responses from different models across various tasks.}
    \label{fig:Percentage of responses}
\end{figure}

\textbf{Comparison with Large-Scale Language Models.} To further demonstrate the effectiveness of our proposed routing framework, we compare it with existing state-of-the-art large-scale LLMs. Herein, we select the $5$ best-performing models in \Cref{tab:base model acc} as experts for the experiment. Specifically, we select a series of well-known open-source LLMs with parameter sizes ranging from 13B to 70B as benchmarks, including Llama-2-13B, Mixtral-8x7B-Instruct-v0.1, Qwen3-30B-A3B, Qwen3-30B-A3B-Instruct-2507, Qwen3-Coder-30B-A3B-Instruct, Qwen2.5-Coder-32B-Instruct and Llama-3.3-70B-Instruct. As shown in \Cref{tab:scale to large model}, our method achieves the highest average score of $83.99\%$, significantly outperforming all individual baseline models. Notably, despite the large parameter count of Llama-3.3-70B-Instruct, our method still surpasses it by an absolute margin of $6.77\%$. Furthermore, our method also demonstrates superior performance in highly specialized domains; for example, it achieves a score of $85.04\%$ on MATH tasks and $77.04\%$ on the more challenging AIME1983-2025 benchmark, far exceeding general-purpose models. By seamlessly coordinating heterogeneous pools of experts, CASLR not only achieves superior performance across both general and highly specialized domains, but also provides an efficient alternative for deploying large-scale LLMs with a large number of parameters (such as 70B). Besides, we also visualize the routing distribution map. As illustrated in \Cref{fig:Percentage of responses}, CASLR can dynamically assign queries to the most specialized experts. For example, the vast majority of Mbpp and AIME queries are assigned to the highly specialized Qwen3-Coder-30B-A3B-Instruct model. This strongly demonstrates that our routing mechanism can effectively utilize the unique strengths of each candidate model in the heterogeneous model pool, effectively avoiding over-reliance on any single general-purpose model.

\begin{table*}[pos=t]
  \centering
  \caption{The impact of the base models' performance on the proposed algorithm.}\label{tab:router quality}%
  \resizebox{\textwidth}{!}{
    \begin{tabular}{c|ccc|cc|cc|c}
    \toprule
    Task  & \multicolumn{3}{c|}{Math} & \multicolumn{2}{c|}{Coding} & \multicolumn{2}{c|}{Reasoning} & \multirow{2}[4]{*}{Average} \\
\cmidrule{1-8}    Models\textbackslash{}Dataset & Math  & GSM-Symbolic & AIME1983-2025 & HumanEval & Mbpp  & MMLU  & HellaSwag &  \\
    \midrule
    \multicolumn{9}{c}{Adding a high-performance model} \\
    \midrule
    Base (1+2+3) & 77.00\% & 92.10\% & 26.02\% & 69.70\% & 61.00\% & 73.76\% & 65.85\% & 66.49\% \\
    Adding DeepSeek-R1-Distill-Llama-8B & 77.76\% & 91.50\% & 54.08\% & 78.79\% & 54.00\% & 73.90\% & 65.70\% & 70.82\% \textcolor{red}{(+4.33\%)} \\
    Adding DeepSeek-R1-Distill-Qwen-14B & 78.44\% & 93.10\% & 64.29\% & 93.94\% & 68.00\% & 82.44\% & 72.05\% & 78.89\% \textcolor{red}{(+12.40\%)}\\
    Adding Llama-3.3-70B-Instruct & 80.60\% & 95.10\% & 55.10\% & 81.82\% & 64.00\% & 86.30\% & 79.25\% & 77.45\% \textcolor{red}{(+10.96\%)}\\
    \midrule
    \multicolumn{9}{c}{Adding a low-performance model} \\
    \midrule
    Base (1+2+3) & 77.00\% & 92.10\% & 26.02\% & 69.70\% & 61.00\% & 73.76\% & 65.85\% & 66.49\% \\
    Adding Llama-2-13B & 76.96\% & 91.80\% & 26.02\% & 57.58\% & 54.00\% & 73.32\% & 65.70\% & 63.63\% \textcolor{green}{(-2.86\%)}\\
    Adding Mixtral-8x7B-Instruct-v0.1 & 76.92\% & 91.40\% & 26.02\% & 51.52\% & 57.00\% & 74.19\% & 65.70\% & 63.25\% \textcolor{green}{(-3.24\%)}\\
    Adding Llama-2-7B & 76.84\% & 92.60\% & 26.02\% & 69.70\% & 56.00\% & 73.43\% & 65.70\% & 65.76\% \textcolor{green}{(-0.73\%)}\\
    \bottomrule
    \end{tabular}}
\end{table*}%

\begin{table*}[pos=t]
  \centering
  \caption{The impact of the number of experts on the proposed algorithm.}\label{tab:num of experts}%
    \resizebox{\textwidth}{!}{
    \begin{tabular}{c|ccc|cc|cc|c}
    \toprule
    Task  & \multicolumn{3}{c|}{Math} & \multicolumn{2}{c|}{Coding} & \multicolumn{2}{c|}{Reasoning} & \multirow{2}[4]{*}{Average} \\
\cmidrule{1-8}    Models\textbackslash{}Dataset & Math  & GSM-Symbolic & AIME1983-2025 & HumanEval & Mbpp  & MMLU  & HellaSwag &  \\
    \midrule
    3 experts (1-3) & 77.00\% & 92.10\% & 26.02\% & 69.70\% & 61.00\% & 73.76\%  & 65.85\% & 66.49\% \\
    4 experts (1-4) & 77.76\% & 91.50\% & 54.08\% & 78.79\% & 54.00\% & 73.90\%  & 65.70\% & 70.82\% \\
    5 experts (1-5) & 81.44\% & 93.20\% & 66.33\% & 93.94\% & 69.00\% & 82.52\%  & 72.10\% & 79.79\% \\
    6 experts (1-6) & 81.52\% & 94.20\% & 67.35\% & 90.91\% & 69.00\% & 82.48\%  & 72.50\% & 79.71\% \\
    7 experts (1-7) & 82.80\% & 96.00\% & 65.82\% & 90.91\% & 64.00\% & 86.88\%  & 79.25\% & 80.81\% \\
    8 experts (1-8) & 83.52\% & 95.50\% & 75.00\% & 78.79\% & 74.00\% & 88.46\%  & 79.05\% & 82.05\% \\
    9 experts (1-9) & 86.80\% & 96.50\% & 81.63\% & 87.88\% & 73.00\% & 89.15\%  & 80.05\% & 85.00\% \\
    10 experts (1-10) & 84.20\% & 95.50\% & 80.10\% & 84.85\% & 72.00\% & 87.89\%  & 79.25\% & 83.40\% \\
    \bottomrule
    \end{tabular}}
\end{table*}%

\subsection{Ablation and Additional Experiments}
\textbf{The impact of adding models with different performance levels on the proposed method.} In this subsection, we explore the impact of expert model quality on the proposed routing framework. To this end, we first construct an expert library containing three models (Qwen2.5-7B-Instruct, Qwen2.5-Coder-7B-Instruct and Qwen2.5-Math-7B-Instruct), denoted as "Base (1+2+3)". Then, we conduct two sets of ablation experiments. In the first set, we introduce a higher-performing model (e.g., DeepSeek-R1-Distill-Llama-8B, DeepSeek-R1-Distill-Qwen-14B or Llama-3.3-70B-Instruct) into the expert library. In the second set, we introduce a lower-performing model (e.g., Llama-2-7B, Llama-2-13B or Mixtral-8x7B-Instruct-v0.1). As shown in \Cref{tab:router quality},  CASLR is highly sensitive to the quality of the candidate model library. Integrating high-performance models can significantly broaden the performance ceiling of the routing framework. Taking the introduction of DeepSeek-R1-Distill-Qwen-14B as an example, the framework not only significantly increases the average accuracy ($66.49\%$) to $78.89\%$, but also achieves breakthroughs in vertical domain performance, resulting in HumanEval and AIME1983-2025 accuracy jumping to $93.94\%$ and $64.29\%$, respectively. In contrast, the introduction of low-performance models has a significant negative effect, causing the overall accuracy of the system to decline significantly, or even fall below the original baseline. For example, the average performance dropped from $66.49\%$ to $63.25\%$ after adding Mixtral-8x7B-Instruct-v0.1. Ultimately, these findings underscore a critical principle for LLM routing: \textit{the quality of the candidate models must be prioritized. An effective routing framework requires strict quality control to prevent poorly performing experts from undermining the benefits of dynamic distribution.}

\textbf{Impact of the Number of Experts.} This experiment aims to evaluate the scalability of the proposed routing algorithm and determine the optimal size of the candidate model library. Specifically, we progressively expand the number of candidate experts from 3 to 10. As observed in \Cref{tab:num of experts}, the average performance is significantly positively correlated with the number of routing experts. This trend indicates that our framework can effectively utilize diverse domain-specific knowledge as the expert library expands. Notably, the performance improvement is most significant in highly specialized tasks, e.g., AIME1983-2025 (from $26.02\%$ to $81.63\%$). However, as the number of models increases further, the marginal gain in performance begins to converge significantly. Although adding more experts (e.g., to 8 or 9) can still bring some performance improvement, this increase gradually slows down. Considering both the marginal benefit of performance improvement and system complexity, we ultimately choose a moderately sized, highly specialized expert pool configuration (i.e., 5 models) as the best practice to balance efficiency and accuracy.

\begin{table*}[pos=t]
  \centering
  \caption{The computational overhead of routers compared to existing LLMs.}\label{tab:time_cost}
  \resizebox{\textwidth}{!}{
    \begin{tabular}{ccccccc}
    \toprule
    Model & DeepSeek-R1-Distill-Qwen-14B & Llama-3.3-70B-Instruct & Qwen3-30B-A3B & Qwen3-30B-A3B-Instruct-2507 & Qwen3-Coder-30B-A3B-Instruct & Router \\
    \midrule
    Time Cost & 3325.10s & 8771.70s & 13724.83s & 9434.59s & 10150.45s & 1.13s \\
    \bottomrule
    \end{tabular}}%
\end{table*}%

\begin{table*}[t]
  \centering
  \caption{Performance comparison between the naive hard-labeling and our soft-labeling mechanism across diverse benchmarks.}\label{tab:soft label vs hard label}
    \resizebox{\textwidth}{!}{
    \begin{tabular}{c|c|ccc|cc|cc|c}
    \toprule
    Task  & \multirow{2}[4]{*}{Type} & \multicolumn{3}{c|}{Math} & \multicolumn{2}{c|}{Coding} & \multicolumn{2}{c|}{Reasoning} & \multirow{2}[4]{*}{Average} \\
\cmidrule{1-1}\cmidrule{3-9}    Models\textbackslash{}Dataset &       & Math  & GSM-Symbolic & AIME1983-2025 & Human-eval & Mbpp  & MMLU  & hellaswag &  \\
    \midrule
    Select the best expert & Hard Label & 78.32\% & 92.10\% & 54.08\% & 87.88\% & 46.00\% & 70.62\% & 70.10\% & 71.30\% \\
    Ours  & Soft Label & 81.44\% & 93.20\% & 66.33\% & 93.94\% & 69.00\% & 82.52\% & 72.10\% & 79.79\% \\
    \bottomrule
    \end{tabular}}%
\end{table*}%

\textbf{The Computational Cost of Router.} To evaluate the computational overhead introduced by our framework, we compare the inference time of the proposed router with the generation time of various candidate LLMs on the HumanEval dataset. As shown in \Cref{tab:time_cost}, our routing mechanism produces extremely low latency ($1.13s$). Ultimately, these results demonstrate that our method not only significantly improves routing accuracy but also operates at a low cost, perfectly meeting the requirements of large-scale practical deployments.



\textbf{Effectiveness of the Soft-Labeling.} To verify our claim that continuous soft labels provide superior supervisory signals compared to one-hot labels, we conduct an ablation study on label construction strategies. We compare our proposed masked softmax method with the naive one-hot baseline discussed in our methodology, which assigns labels to the only successful expert with the highest global utility score in the corresponding semantic cluster of the query. As shown in \Cref{tab:soft label vs hard label}, our soft-labeling mechanism consistently outperforms the hard-labeling strategy across all evaluated benchmarks, yielding an overall improvement of $8.49\%$. These empirical results demonstrate that our soft-labeling mechanism successfully preserves the fine-grained distribution of expertise to establish a highly robust decision boundary.


\section{Conclusion}
This paper identifies and addresses a key bottleneck in existing LLM routing frameworks: routing collapse caused by routing noise. To address this issue, we propose the CASLR framework. CASLR shifts the evaluation paradigm from isolated single-query success rates to macro-domain consensus, leveraging semantic clustering to generate fine-grained soft distributions. These continuous labels effectively filter out routing noise and are used to supervise lightweight BERT routers, enabling them to learn robust and accurate decision boundaries. Extensive experiments on various mathematical, coding, and inference benchmarks demonstrate that CASLR significantly outperforms other competing routing baselines while introducing minimal inference overhead. We hope that CASLR will lay a solid foundation for developing more adaptive LLM routing architectures.

\textbf{Limitations.} While the CASLR framework successfully mitigates routing collapse and achieves superior accuracy, it exhibits certain limitations. Specifically, introducing the newly released expert model requires re-evaluating the global utility score to update the router. Although the amount of computation required for this offline recalculation is small, it still incurs unavoidable costs. Furthermore, the current CASLR framework is limited to a single text modality, making it ill-suited for complex application scenarios involving visual or auditory input.

\textbf{Future Work.} In the future, we aim to extend the capabilities of CASLR from plain text to the multimodal domain. By introducing support for multimodal calls, the router will be able to parse complex cross-modal queries and accurately schedule them to the most suitable multimodal large language model within the heterogeneous ecosystem.










\appendix
\renewcommand{\thefigure}{A\arabic{figure}}
\renewcommand{\thetable}{A\arabic{table}}
\renewcommand{\theHfigure}{appendix.\arabic{figure}}
\renewcommand{\theHtable}{appendix.\arabic{table}}
\setcounter{figure}{0} 
\setcounter{table}{0}  

\section{Introduction to Baseline Methods}
\label{sec:Introduction to Baseline Methods}
In this section, we describe in detail the routing algorithms used in the main text:
\begin{itemize}
    \item \textbf{Oracle}: The maximum possible performance is assumed under the premise that an oracle always selects a single LLM capable of solving each query if it is solvable.~\citep{srivatsa2024harnessing}
    \item \textbf{Random}: This represents the performance of randomly selecting an LLM uniformly for each input query.~\citep{srivatsa2024harnessing}
    \item \textbf{Random\_Oracle\_0.5}~\citep{huang2025routereval}: This simulates a probabilistic routing policy that acts as an oracle (making the optimal selection) with a probability of 0.5, and defaults to a random selection for the remaining queries.
    \item \textbf{MLPR}~\citep{huang2025routereval}: A Multi-Layer Perceptron serves as the classifier.
    \item \textbf{Select the best expert model}: In scenarios where multiple candidate models successfully answer a given query, this strategy resolves the tie by selecting the expert with the highest overall accuracy on the dataset containing that query.
     \item \textbf{PRknn}~\citep{huang2025routereval}: A k-Nearest Neighbors classifier is adopted. During testing, given a test query representation, we retrieve the $K$ train queries with the smallest Euclidean distances. We then compute the average scores of the $m$ models on these $K$ train queries and select the model with the highest average score to process the test query.
    \item \textbf{GraphRouter}~\citep{feng2024graphrouter}: GraphRouter models the performance and cost of LLMs as attributes of edges in the graph and recommends suitable LLMs for specific queries by predicting the characteristics of the edges.
    \item \textbf{C-Bert}~\citep{huang2025routereval}: We use K-Means to cluster the text embeddings of the training set into k clusters. Specifically, for each cluster, the performance of each model in the candidate model pool is evaluated using the samples within that cluster. Then, the best-performing model is selected as the dedicated predictor for that cluster. During inference, the Euclidean distance between the test sample and each cluster center is calculated. The sample is matched to the nearest cluster using a nearest neighbor strategy and the corresponding dedicated predictor is used as the router result.
    \item \textbf{LinearR}~\citep{huang2025routereval}: A simple linear layer is used as the classifier. The input to the linear layer is the query representation, and the output dimension corresponds to the number of LLMs in the candidate pool, representing the selection scores for each LLM.
    \item \textbf{MLC}~\citep{huang2025routereval}: MLC treats multiple labels as positive classes for supervised learning. During inference, the input embedding passes through the classifier to generate a probability distribution, and selects the highest one as the final router result.
\end{itemize}

\begin{table*}[pos=t]
  \centering
  \caption{Performance of the base models.}
  \resizebox{\textwidth}{!}{
    \begin{tabular}{c|c|cc|cc|ccc|c}
    \toprule
    \multirow{2}[4]{*}{ID} & Task  & \multicolumn{2}{c|}{Math} & \multicolumn{2}{c|}{Coding} & \multicolumn{3}{c|}{Reasoning} & \multirow{2}[4]{*}{Average} \\
\cmidrule{2-9}          & Models\textbackslash{}Dataset & Math  & GSM-Symbolic & HumanEval & Mbpp  & MMLU  & AIME1983-2025 & HellaSwag &  \\
    \midrule
    1     & Qwen2.5-7B-Instruct & 71.56\% & 87.40\% & 69.70\% & 54.00\% & 73.54\% & 23.47\% & 65.70\% & 63.62\% \\
    2     & Qwen2.5-Coder-7B-Instruct & 64.40\% & 74.80\% & 75.76\% & 61.00\% & 66.11\% & 14.80\% & 57.95\% & 59.26\% \\
    3     & Qwen2.5-Math-7B-Instruct & 76.92\% & 89.80\% & 30.30\% & 35.00\% & 52.70\% & 26.02\% & 39.80\% & 50.08\% \\
    4     & DeepSeek-R1-Distill-Llama-8B & 64.16\% & 71.60\% & 81.82\% & 50.00\% & 71.41\% & 68.37\% & 59.35\% & 66.67\% \\
    5     & DeepSeek-R1-Distill-Qwen-14B & 71.68\% & 84.90\% & 93.94\% & 68.00\% & 82.66\% & 67.35\% & 72.05\% & 77.23\% \\
    6     & Llama-3.1-8B-Instruct & 55.72\% & 79.60\% & 66.67\% & 51.00\% & 71.99\% & 11.22\% & 64.70\% & 57.27\% \\
    7     & Llama-3.3-70B-Instruct & 80.60\% & 93.50\% & 81.82\% & 64.00\% & 86.30\% & 55.10\% & 79.25\% & 77.22\% \\
    8     & Qwen3-30B-A3B & 72.76\% & 74.20\% & 69.70\% & 71.00\% & 85.29\% & 79.59\% & 74.75\% & 75.33\% \\
    9     & Qwen3-30B-A3B-Instruct-2507 & 83.24\% & 76.40\% & 90.91\% & 71.00\% & 85.18\% & 81.63\% & 78.15\% & 80.93\% \\
    10    & Mixtral-8x7B-Instruct-v0.1 & 23.32\% & 62.80\% & 54.55\% & 41.00\% & 68.89\% & 1.53\% & 63.10\% & 45.03\% \\
    11    & Qwen3-Coder-30B-A3B-Instruct & 82.20\% & 91.10\% & 90.91\% & 71.00\% & 82.05\% & 49.49\% & 71.40\% & 76.88\% \\
    \bottomrule
    \end{tabular}}
  \label{tab:base model acc}%
\end{table*}%

\begin{figure}
    \centering
    \includegraphics[width=0.49\textwidth]{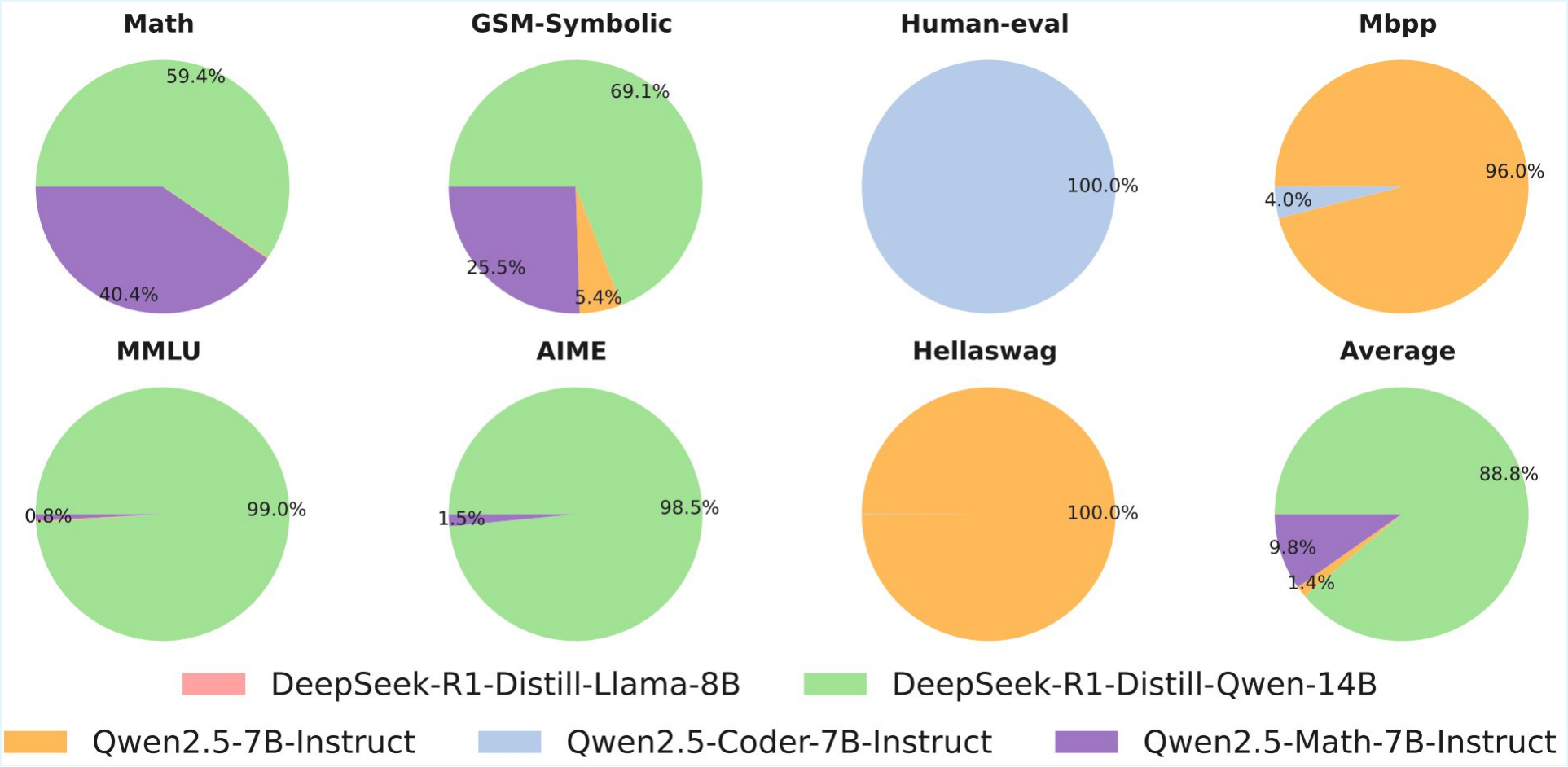}
    \caption{Percentage of responses from different models across various tasks.}
    \label{fig:task_composition1}
\end{figure}

\begin{figure}
    \centering
    \includegraphics[width=0.49\textwidth]{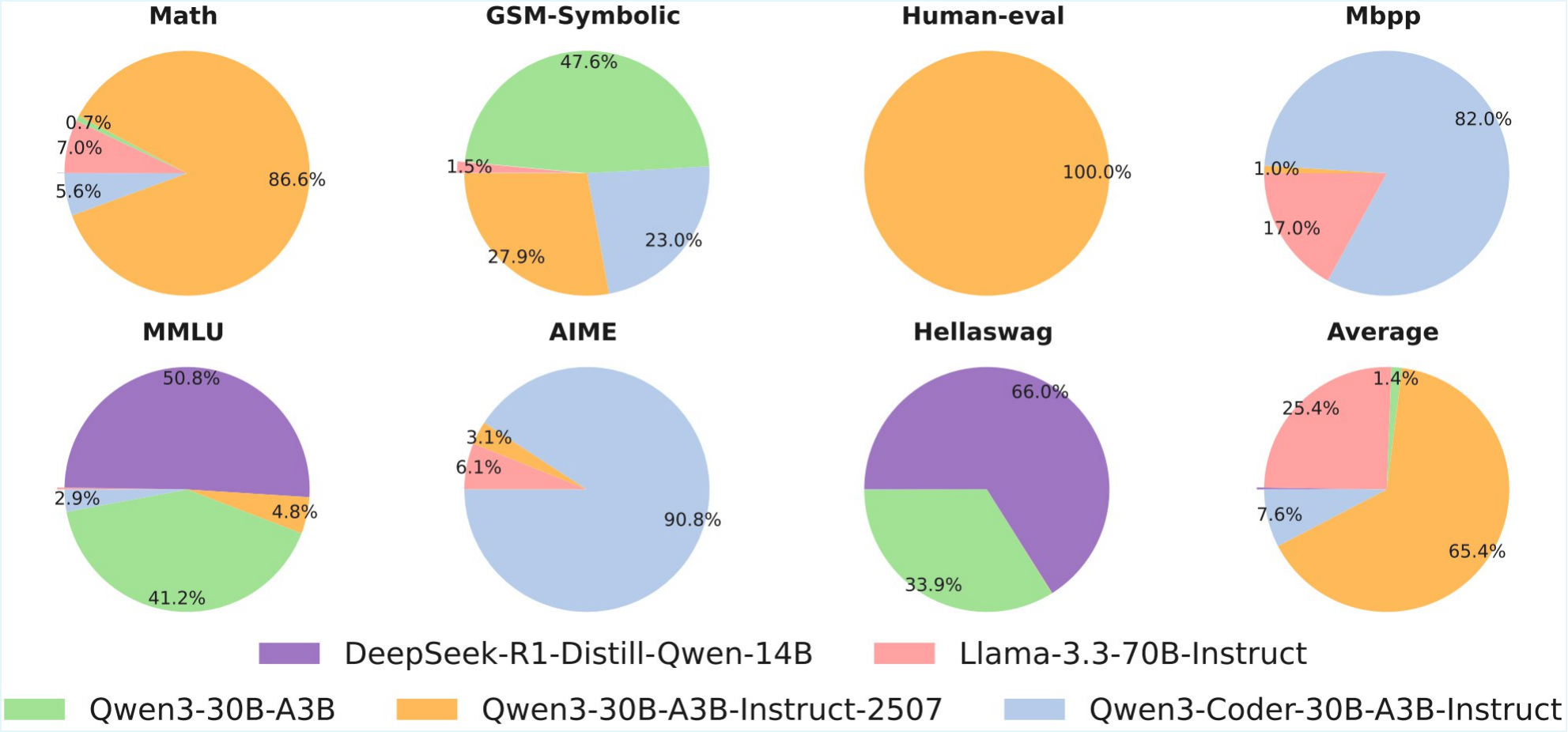}
    \caption{Percentage of responses from different models across various tasks.}
    \label{fig:task_composition2}
\end{figure}

\begin{figure}
    \centering
    \includegraphics[width=0.49\textwidth]{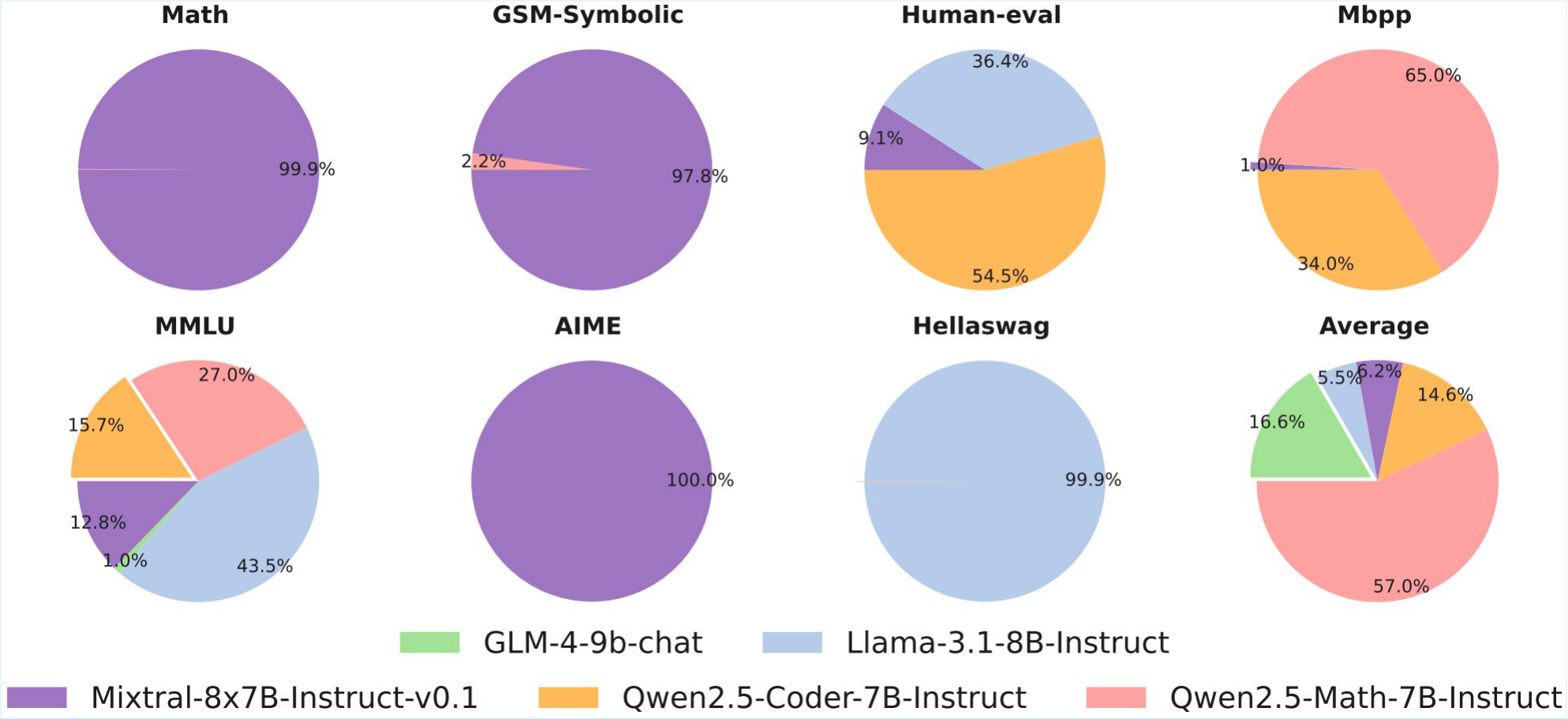}
    \caption{Percentage of responses from different models across various tasks.}
    \label{fig:task_composition3}
\end{figure}

\begin{figure}
    \centering
    \includegraphics[width=0.49\textwidth]{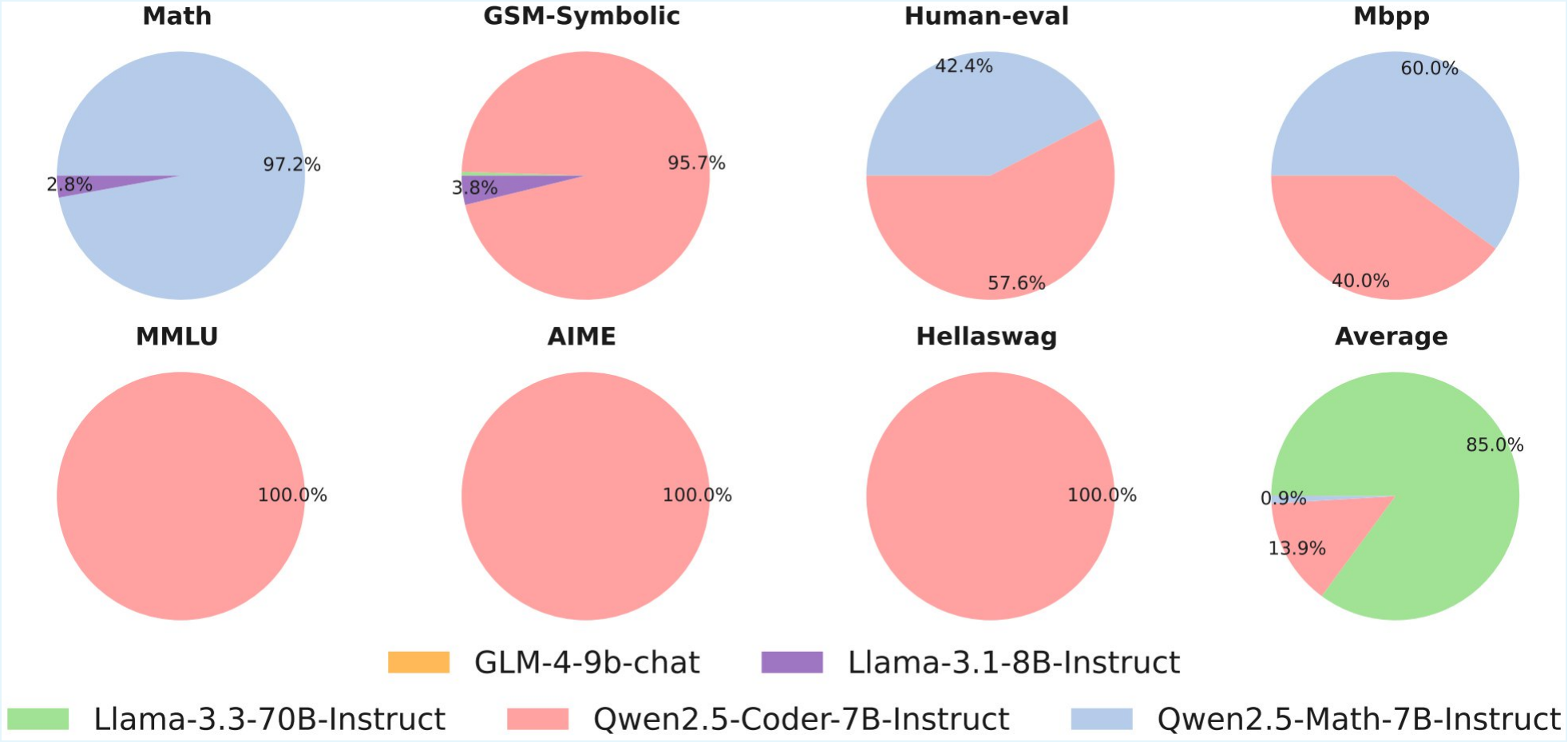}
    \caption{Percentage of responses from different models across various tasks.}
    \label{fig:task_composition4}
\end{figure}

\section{Declaration of Competing Interest}
The authors declare that they have no known competing financial interests or personal relationships that could have appeared to influence the work reported in this paper.


\printcredits

\bibliographystyle{cas-model2-names}

\bibliography{cas-refs}



\end{document}